\documentclass[11pt,a4paper]{article}
\usepackage[hyperref]{acl2021}
\usepackage{times}
\usepackage{latexsym}

\usepackage{microtype}
\usepackage{float}
\usepackage{helvet} 
\usepackage{courier}  
\usepackage{url}  
\usepackage{graphicx} 
\usepackage{natbib}  
\usepackage{booktabs} 
\usepackage{multirow}
\usepackage{siunitx}
\usepackage{colortbl}
\everypar{\looseness=-1} 
\usepackage{subcaption}
\usepackage{amsmath}

\usepackage{amsmath}
\usepackage{amsfonts}
\usepackage{breqn}

\aclfinalcopy

\newcommand\Tstrut{\rule{0pt}{2.0ex}}         
\newcommand\Bstrut{\rule[-0.9ex]{0pt}{0pt}}   

\makeatletter
\newcommand*\dotproduct{\mathpalette\dotproduct@{.5}}
\newcommand*\dotproduct@[2]{\mathbin{\vcenter{\hbox{\scalebox{#2}{$\m@th#1\bullet$}}}}}
\makeatother

\title{Learning a Reversible Embedding Mapping using Bi-Directional Manifold Alignment}

\author{Ashwinkumar Ganesan\thanks{This research was completed prior to joining Amazon.}, Francis Ferraro, Tim Oates \\
  Dept. Of Computer Science \& Electrical Engineering (CSEE), \\
  University Of Maryland Baltimore County (UMBC), \\
  MD, USA - 21250 \\
  {\tt gashwin1@umbc.edu, ferraro@umbc.edu, oates@cs.umbc.edu} \\}

\begin{document}
\maketitle
\begin{abstract}
We propose a Bi-Directional Manifold Alignment (BDMA) that learns a non-linear mapping between two manifolds by explicitly training it to be bijective. We demonstrate BDMA by training a model for a pair of languages rather than individual, directed source and target combinations, reducing the number of models by $50$\%. We show that models trained with BDMA in the ``forward'' (source to target) direction can successfully map words in the ``reverse'' (target to source) direction, yielding equivalent (or better) performance to standard unidirectional translation models where the source and target language is flipped. We also show how BDMA reduces the overall size of the model.
\end{abstract}

\section{Introduction}
Learning continuous vector representations of embeddings is an expensive exercise as it requires a large quantity of free text to train \textit{stable} representations \citep{sahin2017consistent}. Learning word embeddings in the English language is relatively easy since a model can make use of free text online from sources like \textit{Wikipedia}, but it is challenging to learn embeddings for natural languages where the free text is limited (low-resource languages). Resource-constrained languages suffer from dual problems of reduced quality of embeddings and their vocabulary being small. Cross-lingual words embedding (CLWE) models alleviate this problem but are often linear mapping functions that align the source and target language manifolds, since non-linear mapping functions such as neural networks are unidirectional and known to perform poorly as compared to their linear counterparts \citep{ruder2019survey}. 

In this paper, we propose Bi-Directional Manifold Alignment (\textbf{BDMA}), which learns a \textit{reversible}, non-linear mapping function between two manifolds. Inspired by CycleGAN \citep{zhu2017unpaired}, we use a \textit{cycle consistency} loss to optimize BDMA. We study BDMA in the context of cross-lingual lexicon induction and show that it offers solutions to two known problems: (1) that non-linear models are known to perform poorly in comparison to their linear counterparts \cite{ruder2019survey}, and (2) most approaches perform unidirectional mapping only (from a source to target language), leading to an ever increasing set of translation models. We show how BDMA is a \textit{generic} training method that uses different distance metrics (or losses) like MSE, cosine or RCSLS \cite{joulin2018loss} while training models cyclically.\footnote{Implementation see \href{https://github.com/codehacken/bdma}{https://github.com/codehacken/bdma}.}


\begin{figure}[t]
    \centering
    \includegraphics[scale=0.45]{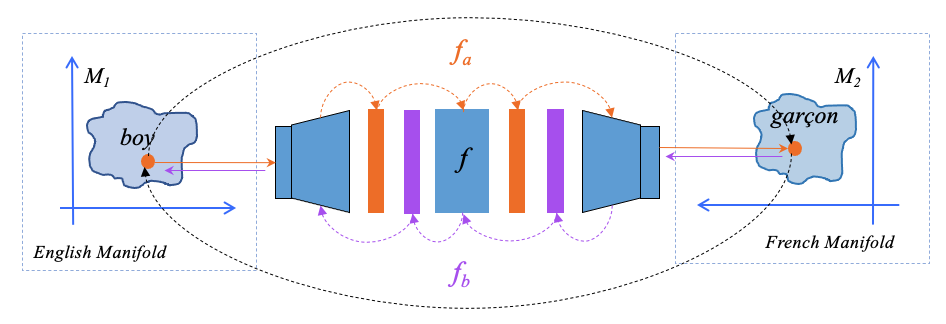}
\caption{\textbf{Mapping vector spaces with Bi-Directional Manifold Alignment (BDMA).} $f$ is the feedforward network. $f_a$ and $f_b$ are represent the forward and backward direction of flow through the network. In a shared BDMA network, the {\color{blue}blue} components represent network fully connected layers, {\color{orange}orange} are activation layers during {\color{orange}forward network flow} while {\color{violet}purple} represents activation layers in {\color{violet}reverse flow}. During reverse flow from output to input, the weight matrix is a transpose of weights during forward flow through the network.}
    \label{fig:align_ccl_lpp}
\end{figure}

\section{Bi-Directional Manifold Alignment}
Consider two manifolds $M^s\in\mathbb{R}^{n\times d}$ (source domain) and $M^t\in\mathbb{R}^{m\times d}$ (target domain) that are vector space representations of words. The monolingual word embeddings are pretrained from a large corpus and may be created using different methods. Let $V^s$ and $V^t$ be the respective vocabularies of the two languages. Hence $V^s = \{w^s_1\ ...\ w^s_n\}$ and $V^t = \{w^t_1\ ..\ w^t_m\}$ are words in each vocabulary of size $n$ and $m$. The distributed representations of words in each manifold are $M^s = \{m^s_1\ ...\ m^s_n\}$ and $M^t = \{m^t_1\ ...\ m^t_m\}$. We assume there is $V^p = \{w^p_1\ ...\ w^p_c\}$, an available dictionary or parallel corpus of words for the given source/target pair.

\subsection{Bi-Directional Loss Mechanism} 
\label{sec:bi-loss}
We achieve bi-directional alignment by learning a mapping function is optimized with a \textit{cyclic-consistency loss} (CCL). In Figure \ref{fig:align_ccl_lpp}, the mapping function $f_a: M^s \rightarrow M^t$ to align the manifold $M^s$ to $M^t$. We also use a backward mapping function $f_b: M^t \rightarrow M^s$ to align the manifold $M^t$ to $M^s$. We refer to the parameters of both $f_a$ and $f_b$ as $\theta_f$. 

Our method is based on jointly minimizing the distance $\mathcal{D}$ between pairs of embeddings, and their mapped counterparts, from each manifold. %
We define our cycle consistency loss for a single training sample based on this distance function $\mathcal{D}$ as %
\begin{equation}
    \mathcal{L}_{\textrm{ccl}}^{\mathcal{D}}(i) = \mathcal{D}(f_a(m^s_i), m^t_i) + \mathcal{D}(f_b(m^t_i), m^s_i).
    \label{eqn:l_ccl}
\end{equation}

\noindent Following previous work~\cite{xing2015normalized}, we include an \textit{orthogonal loss} in the objective; we extend this loss function for a neural network by performing a \textbf{layerwise} orthogonal loss. For our full objective, we sum over all training instances and minimize over $\theta_f$:
\begin{equation}
    \mathcal{L}_{\textrm{ccl}} = 
    \min_{\theta_f} \sum_{i \in V^p} \mathcal{L}_{\textrm{ccl}}^{\mathcal{D}}(i) + \sum_{w_j \in \theta_f} w_jw_j^T - I
\end{equation}
where $w_j$ are weights of layer $j$ in the network. While Euclidean distance (mean squared error: $\mathcal{D}=\mathrm{MSE}$) is a common way of computing distance in a manifold~\cite{ruder2019survey,artetxe2016learning}, cosine or relaxed cross-domain similarity local scaling (RCSLS) \cite{joulin2018loss} distance functions have been shown to be effective for word and embedding alignment tasks. Our formulation works with these other computable distance functions. For example, while applying $\mathcal{D}={\textrm{MSE}}$, and for ease omitting the orthogonal loss term $\sum_{w_j \in \theta_f} w_jw_j^T - I$, the loss is
\begin{equation*}
\small
\min_{\theta_f} \sum_{i \in V^p} \left\|f_a(m^s_i) - m^t_i\right\|^2_2 + \left\|f_b(m^t_i) - m^s_i\right\|^2_2.
\label{eqn:l_mse}
\end{equation*}
See Appendix \ref{sec:app_additional_loss} for similar formulations for $\mathcal{D} = \mathrm{cosine}$, $\mathcal{D} = \mathrm{RCSLS}$, and a combined distance function $\mathcal{D} = \mathrm{cosine+RCSLS}$ (used in \S \ref{subsec:exp_muse_dataset}).

\subsection{Forward - Reverse Network Flow}
As described in \S \ref{sec:bi-loss}, $f_a$ and $f_b$ represent the forward and reverse network flow. We represent the forward and reverse mapping with two networks that have shared or independent parameters. When the parameters are independent, two separate networks are trained simultaneously and optimized in order to learn the mapping between two languages. In Figure \ref{fig:align_ccl_lpp}, the network parameters are shared in our model. The forward flow is shown in {\color{orange}orange} while reverse flow is depicted in {\color{violet}purple}.

Although the two networks share parameters, they cannot do so directly as the required shapes of each layer differ. In order to perform backward translation, reverse flow is enabled in the network by explicitly taking the \textbf{transpose of each layer} in the network (we use fully connected layers without bias vectors) making the network bi-directional or invertible. With our cycle consistency loss formulation, the model learns layers such that the transpose of the layer inverts the network.

\begin{table*}
  \centering
    \resizebox{\textwidth}{!}{
    \begin{tabular}{c | c | c | c | c | c | c | c | c }
        \toprule
        \multirow{2}{*}{\textbf{Method}} & \multicolumn{6}{c}{\textbf{Evaluation [\textit{s} $\rightarrow$ \textit{t}] P@1 }}\Tstrut\Bstrut\\\cline{2-9}
         & \textbf{En}$\rightarrow$\textbf{Es}  & \textbf{Es}$\rightarrow$\textbf{En}  & \textbf{En}$\rightarrow$\textbf{It}  & \textbf{It}$\rightarrow$\textbf{En} & \textbf{En}$\rightarrow$\textbf{De}  & \textbf{De}$\rightarrow$\textbf{En} & \textbf{En}$\rightarrow$\textbf{Fr}  & \textbf{Fr}$\rightarrow$\textbf{En} \Tstrut\Bstrut\\
        \hline
        MUSE \cite{conneau2017word}\textbf{*} & 81.7 & 83.3 & 77.7 & 78.2 & 74.0 & 72.0 & 82.3 & 82.1 \Tstrut\Bstrut\\
        VECMAP \cite{artetxe2018generalizing} & 80.80 & 85.20 & 77.47 & 80.47 & 73.33 & 75.07 & 81.60 & 84.40 \Tstrut\Bstrut\\
        GeoMM \cite{jawanpuria2019learning} & 81.53 & 86.33 & 78.47 & 81.53 & 74.80 & 76.67 & 82.00 & 84.67 \Tstrut\Bstrut\\
    	RCSLS \cite{joulin2018loss}\textbf{*} & 84.1 & 86.3 & 78.5 & 79.8 & \textbf{79.1} & 76.3 & 83.3 & 84.1 \Tstrut\Bstrut\\
    	\hline
    	BLISS(R) \cite{patra2019bilingual}\textbf{*} & \textbf{84.3} & 86.2 & \textbf{79.3} & 82.4& \textbf{79.1} & 76.6 & 83.9 & 84.7 \Tstrut\Bstrut\\
    	Joint Align \cite{wang2019cross}\textbf{**} & 69.6 & 71.9 & - & - & 68.7 & 70.7 & 78.0 & 79.2 \Tstrut\Bstrut\\
    	Cross-lingual Anchoring \cite{ormazabal2020beyond}\textbf{**} & 84.2 & 86.5 & - & - & 78.1 & 76.9 & \textbf{84.9} & \textbf{85.0} \Tstrut\Bstrut\\
    	LNMAP (LIN. AE) \cite{mohiuddin2020lnmap}\textbf{*} & 82.9 & 86.4 & 78.1 & 81.4 & 75.5 & 75.9 & 83.9 & 84.7 \Tstrut\Bstrut\\
    	\hline
    	\multicolumn{8}{c}{\textit{\textbf{Linear Mapping}}}\Tstrut\Bstrut\\
        \hline
        BDMA \texttt{[C + R]} (\textit{s} $\rightarrow$ \textit{t}) & \cellcolor{blue!25}83.13 & 83.26 & \cellcolor{blue!25}78.60 & 78.60 & \cellcolor{blue!25}76.13 & 74.73 & \cellcolor{blue!25}83.73 & 82.86 \Tstrut\Bstrut\\
        BDMA \texttt{[C + R]} (\textit{t} $\rightarrow$ \textit{s}) & 83.13 & \cellcolor{blue!25}84.06 & 78.60 & \cellcolor{blue!25}78.53 & 73.46 & \cellcolor{blue!25}75.66 & 82.8 & \cellcolor{blue!25}83.86 \Tstrut\Bstrut\\
        \hline
        \multicolumn{8}{c}{\textit{\textbf{1-Hidden Layer Feedforward Network}}}\Tstrut\Bstrut\\
        \hline
        BDMA \texttt{[C + R]} (\textit{s} $\rightarrow$ \textit{t}) & \cellcolor{blue!25}82.40 & 85.73 & \cellcolor{blue!25}78.66 & 82.60 & \cellcolor{blue!25}74.46 & \textbf{78.40} & \cellcolor{blue!25}83.40 & 84.93 \Tstrut\Bstrut\\
        BDMA \texttt{[C + R]} (\textit{t} $\rightarrow$ \textit{s}) & 81.60 & \cellcolor{blue!25}\textbf{86.80} & 78.4 & \cellcolor{blue!25}\textbf{82.66} & 73.46 & \cellcolor{blue!25}74.86 & 79.86 & \cellcolor{blue!25}84.33 \Tstrut\Bstrut\\
        \bottomrule
    \end{tabular}
  }
  \caption{\textbf{Bi-Directional Manifold Alignment (BDMA) measured with Precision @ 1} measured on filtered MUSE evaluation set \textbf{(with polysemous words) for high resource languages}. \textbf{*} represents results taken directly from the cited paper. \textbf{**} represents results taken from \citet{ormazabal2020beyond}. We consider the \textit{best} results for each language pair and direction from this paper. \textbf{-} represents language pairs that are not part of experiments in the original paper. BLISS(R) \cite{patra2019bilingual} is semi-supervised.}
  \label{tab:exp_base_hr_model_performance_combined}
\end{table*}

\begin{table*}[h!]
   \centering
    \resizebox{\textwidth}{!}{
    \begin{tabular}{c | c | c | c | c | c | c | c | c }
        \toprule
        \multirow{2}{*}{\textbf{Method}} & \multicolumn{6}{c}{\textbf{Evaluation [\textit{s} $\rightarrow$ \textit{t}] P@1}}\Tstrut\Bstrut\\\cline{2-9}
         & \textbf{En}$\rightarrow$\textbf{Ru}  & \textbf{Ru}$\rightarrow$\textbf{En}  & \textbf{En}$\rightarrow$\textbf{Hi}  & \textbf{Hi}$\rightarrow$\textbf{En} & \textbf{En}$\rightarrow$\textbf{Ja}  & \textbf{Ja}$\rightarrow$\textbf{En} & \textbf{En}$\rightarrow$\textbf{Pt}  & \textbf{Pt}$\rightarrow$\textbf{En} \Tstrut\Bstrut\\
        
        \hline
        VECMAP \cite{artetxe2016learning} & 52.33 & 65.73 & 34.87 & 50.03 & 51.54 & 41.42 & 80.27 & 80.67 \Tstrut\Bstrut\\
        VECMAP \cite{artetxe2018generalizing} & 51.53 & 70.00 & \textbf{40.40} & \textbf{56.46} & 46.95 & 44.25 & 80.60 & 82.93 \Tstrut\Bstrut\\
        GeoMM \cite{jawanpuria2019learning} & 54.13 & 69.47 & - & 54.72 & 27.55 & 23.66 & \textbf{81.60} & 83.27 \Tstrut\Bstrut\\
    	
    	\hline
    	\multicolumn{8}{c}{\textit{\textbf{Linear Mapping}}}\Tstrut\Bstrut\\
        \hline
        BDMA \texttt{[C + R]} (\textit{s} $\rightarrow$ \textit{t}) & \cellcolor{blue!25}55.80 & 68.66 & \cellcolor{blue!25}36.80 & 54.58 & \cellcolor{blue!25}53.59 & 38.52 & \cellcolor{blue!25}80.40 & \textbf{84.20} \Tstrut\Bstrut\\
        BDMA \texttt{[C + R]} (\textit{t} $\rightarrow$ \textit{s}) & 55.60 & \cellcolor{blue!25}69.73 & 36.60 & \cellcolor{blue!25}55.25 & 53.52 & \cellcolor{blue!25}38.73 & 80.06 & \cellcolor{blue!25}83.93 \Tstrut\Bstrut\\
        \hline
        \multicolumn{8}{c}{\textit{\textbf{1-Hidden Layer Feedforward Network}}}\Tstrut\Bstrut\\
        \hline
        BDMA \texttt{[C + R]} (\textit{s} $\rightarrow$ \textit{t}) & \cellcolor{blue!25}\textbf{57.20} & \textbf{70.20} & \cellcolor{blue!25}37.00 & 54.11 & \cellcolor{blue!25}\textbf{54.07} & 46.51 & \cellcolor{blue!25}80.13 & 83.13 \Tstrut\Bstrut\\
        BDMA \texttt{[C + R]} (\textit{t} $\rightarrow$ \textit{s}) & 56.93 & \cellcolor{blue!25}70.06 & 37.00 & \cellcolor{blue!25}54.38 & 54.28 & \cellcolor{blue!25}\textbf{47.07} & 80.06 & \cellcolor{blue!25}83.93 \Tstrut\Bstrut\\        
        \bottomrule
    \end{tabular}
  }
  \caption{\textbf{Bi-Directional Manifold Alignment (BDMA) measured with Precision @ 1} on default MUSE evaluation set \textbf{(with polysemous words) for \textbf{low-resource languages}}.}
  \label{tab:exp_base_lr_model_performance_combined}
\end{table*}

\section{Experiments \& Analysis}
\label{subsec:exp_muse_dataset}
We experiment with the \textbf{MUSE} dataset \cite{conneau2017word}. It consists of $110$ bilingual dictionaries with separate training and test datasets for each language pair. The pairs contain polysemous words. When it comes to training BDMA, polysemous words can provide additional context to the model being trained while handicapping other baseline models. We filter out training pairs for polysemous words (source or target). The models are trained with $5000$ unique pairs. We show two sets of experiments: (a) with a filtered evaluation set that contains $1500$ unique pairs and (b) with the original evaluation dataset. We measure the performance of BDMA on two sets of languages: the low-resource languages \textit{Russian} (Ru) and \textit{Japanese} (Ja), and the high-resource languages \textit{Spanish} (Es), \textit{French} (Fr), \textit{German} (De) and \textit{Italian} (It).

In each table, \textbf{\textit{s}} is the source language while \textbf{\textit{t}} is the target language. $\rightarrow$ indicates the direction of mapping and training language pairs used from MUSE. For \textbf{reverse translation}, the model is trained with the \textit{t} $\rightarrow$ \textit{s} dataset and evaluated on the \textit{s} $\rightarrow$ \textit{t} test dataset---for example, the model trained on \textbf{En}$\rightarrow$\textbf{Ru} is evaluated on \textbf{Ru}$\rightarrow$\textbf{En}. P@1 measurements highlighted in \textbf{{\color{blue}blue show the forward (training) direction}} in which the model is trained and its adjacent non-colored measurement uses the \textbf{same model} to perform \textbf{reverse} translation.

\begin{table*}
    \centering
    \resizebox{\textwidth}{!}{
    \begin{tabular}{c | c | c | c | c | c | c | c | c }
        \toprule
        \multirow{2}{*}{\textbf{Method}} & \multicolumn{6}{c}{\textbf{Evaluation [\textit{s} $\rightarrow$ \textit{t}] P@1 }}\Tstrut\Bstrut\\\cline{2-9}
        & \textbf{En}$\rightarrow$\textbf{Es}  & \textbf{Es}$\rightarrow$\textbf{En}  & \textbf{En}$\rightarrow$\textbf{It}  & \textbf{It}$\rightarrow$\textbf{En} & \textbf{En}$\rightarrow$\textbf{De}  & \textbf{De}$\rightarrow$\textbf{En} & \textbf{En}$\rightarrow$\textbf{Fr}  & \textbf{Fr}$\rightarrow$\textbf{En} \Tstrut\Bstrut\\
        
        \hline
        MUSE \cite{conneau2017word} & 48.06 & 61.27 & 51.33 & 62.59 & 37.4 & 50.21 & 39.33 & 51.60 \Tstrut\Bstrut\\
        VECMAP \cite{artetxe2016learning} & 48.87 & 61.49 & 52.07 & 62.31 & 38.60 & 50.22 & 47.47 & 59.10\Tstrut\Bstrut\\
        VECMAP \cite{artetxe2018generalizing} & 48.27 & 62.79 & 52.20 & 65.24 & 37.80 & 52.59 & 47.67 & 60.89 \Tstrut\Bstrut\\
        GeoMM \cite{jawanpuria2019learning} & 48.60 & 63.79 & 52.53 & 65.38 & 38.33 & 53.45 & 48.60 & \textbf{61.24} \Tstrut\Bstrut\\
    	RCSLS \cite{joulin2018loss} & 49.26 & \textbf{64.29} & \textbf{53.00} & \textbf{66.52} & 38.93 & \textbf{53.73} & 47.66 & 59.74 \Tstrut\Bstrut\\
    	\hline
    	\multicolumn{8}{c}{\textit{\textbf{Linear Mapping}}}\Tstrut\Bstrut\\
        \hline
        BDMA \texttt{[C + R]} (\textit{s} $\rightarrow$ \textit{t}) & \cellcolor{blue!25}\textbf{49.40} & 62.57 & \cellcolor{blue!25}52.80 & 63.09 & \cellcolor{blue!25}\textbf{39.33} & 52.37 & \cellcolor{blue!25}48.73 & 59.95 \Tstrut\Bstrut\\
        BDMA \texttt{[C + R]} (\textit{t} $\rightarrow$ \textit{s}) & 49.33 & \cellcolor{blue!25}62.78 & 52.46 & \cellcolor{blue!25}63.31 & 39.00 & \cellcolor{blue!25}52.01 & 49.06 & \cellcolor{blue!25}59.38 \Tstrut\Bstrut\\
        \hline
        \multicolumn{8}{c}{\textit{\textbf{1-Hidden Layer Feedforward Network}}}\Tstrut\Bstrut\\
        \hline
        BDMA \texttt{[C + R]} (\textit{s} $\rightarrow$ \textit{t}) & \cellcolor{blue!25}48.90 & 62.28 & \cellcolor{blue!25}52.06 & 65.02 & \cellcolor{blue!25}38.46 & 51.86 & \cellcolor{blue!25}\textbf{48.86} & 60.67 \Tstrut\Bstrut\\
        BDMA \texttt{[C + R]} (\textit{t} $\rightarrow$ \textit{s}) & 48.46 & \cellcolor{blue!25}63.00 & 52.46 & \cellcolor{blue!25}65.09 & 39.06 & \cellcolor{blue!25}52.08 & 47.33 & \cellcolor{blue!25}60.67 \Tstrut\Bstrut\\
        \bottomrule
    \end{tabular}
  }    
  \caption{\textbf{Bi-Directional Manifold Alignment (BDMA) measured with Precision @ 1} shows the performance of different models on \textbf{high-resource languages} in the MUSE dataset \cite{conneau2017word} in comparison to \textit{BDMA}. The test dataset contains \textbf{unique pairs} only.}
  \label{tab:exp_base_hr_model_performance}
\end{table*}

\begin{table*}[h!]
  \centering
    \resizebox{\textwidth}{!}{
    \begin{tabular}{c | c | c | c | c | c | c | c | c }
        \toprule
        \multirow{2}{*}{\textbf{Method}} & \multicolumn{6}{c}{\textbf{Evaluation [\textit{s} $\rightarrow$ \textit{t}]}}\Tstrut\Bstrut\\\cline{2-9}
         & \textbf{En}$\rightarrow$\textbf{Ru}  & \textbf{Ru}$\rightarrow$\textbf{En}  & \textbf{En}$\rightarrow$\textbf{Hi}  & \textbf{Hi}$\rightarrow$\textbf{En} & \textbf{En}$\rightarrow$\textbf{Ja}  & \textbf{Ja}$\rightarrow$\textbf{En} & \textbf{En}$\rightarrow$\textbf{Pt}  & \textbf{Pt}$\rightarrow$\textbf{En} \Tstrut\Bstrut\\
        \hline
        VECMAP \cite{artetxe2016learning} & 35.27 & 52.28 & 23.80 & 26.45 & 39.07 & 35.59 & 44.93 & 61.06 \Tstrut\Bstrut\\
        VECMAP \cite{artetxe2018generalizing} & 34.53 & 56.28 & 27.40 & \textbf{31.21} & 36.20 & 38.88 & 49.53 & 63.83 \Tstrut\Bstrut\\
        GeoMM \cite{jawanpuria2019learning} & 36.93 & \textbf{56.42} & \textbf{27.67} & 30.35 & 21.60 & 21.81 & \textbf{50.13} & 63.90 \Tstrut\Bstrut\\
    	RCSLS \cite{joulin2018loss} & 37.73 & 54.85 & 24.80 & 26.80 & 39.40 & \textbf{38.80} & 49.06 & \textbf{64.53} \Tstrut\Bstrut\\
    	
    	\hline
    	\multicolumn{8}{c}{\textit{\textbf{Linear Mapping}}}\Tstrut\Bstrut\\
        \hline
        BDMA \texttt{[C + R]} (\textit{s} $\rightarrow$ \textit{t}) & \cellcolor{blue!25}37.40 & 52.35 & \cellcolor{blue!25}25.20 & 26.87 & \cellcolor{blue!25}40.8 & 34.64 & \cellcolor{blue!25}49.46 & 63.12 \Tstrut\Bstrut\\
        BDMA \texttt{[C + R]} (\textit{t} $\rightarrow$ \textit{s}) & 36.93 & \cellcolor{blue!25}52.99 & 24.73 & \cellcolor{blue!25}27.37 & 40.4 & \cellcolor{blue!25}35.30 & 49.00 & \cellcolor{blue!25}63.40
        \Tstrut\Bstrut\\
        \hline
        \multicolumn{8}{c}{\textit{\textbf{1-Hidden Layer Feedforward Network}}}\Tstrut\Bstrut\\
        \hline
        BDMA \texttt{[C + R]} (\textit{s} $\rightarrow$ \textit{t}) & \cellcolor{blue!25}37.73 & 52.06 & \cellcolor{blue!25}25.13 & 27.73 & \cellcolor{blue!25}\textbf{40.93} & 37.70 & \cellcolor{blue!25}49.46 & 63.12 \Tstrut\Bstrut\\
        BDMA \texttt{[C + R]} (\textit{t} $\rightarrow$ \textit{s}) & \textbf{38.40} & \cellcolor{blue!25}53.49 & 25.13 & \cellcolor{blue!25}29.14 & 40.00 & \cellcolor{blue!25}38.14 & 48.73 & \cellcolor{blue!25}64.32
        \Tstrut\Bstrut\\        
        \bottomrule
    \end{tabular}
  }
  \caption{\textbf{Bi-Directional Manifold Alignment (BDMA) measured with Precision @ 1} on unique MUSE evaluation set \textbf{(without polysemous words)} for \textbf{low-resource languages}.}
  \label{tab:exp_base_lr_model_performance_default}
\end{table*}

\begin{table}[t]
  \centering
  \resizebox{\columnwidth}{!}{
    \begin{tabular}{c | c | c | c | c }
        \toprule
        \multirow{2}{*}{\textbf{Loss}} & \multicolumn{3}{c}{\textbf{Evaluation [\textit{s} $\rightarrow$ \textit{t}] P@1 }}\Tstrut\Bstrut\\\cline{2-5}
         & \textbf{En}$\rightarrow$\textbf{Ru}  & \textbf{Ru}$\rightarrow$\textbf{En}  & \textbf{En}$\rightarrow$\textbf{Ja}  & \textbf{Ja}$\rightarrow$\textbf{En}\Tstrut\Bstrut\\
        \hline
    	\multicolumn{5}{c}{\textit{\textbf{Linear Mapping}} (\textit{s} $\rightarrow$ \textit{t})}\Tstrut\Bstrut\\
        \hline
        \texttt{[M]}  & \cellcolor{blue!25}54.66 & 66.26 & \cellcolor{blue!25}21.93 & 16.26 \Tstrut\Bstrut\\
        \texttt{[C]}  & \cellcolor{blue!25}55.00 & 66.46  & \cellcolor{blue!25}52.22 & 40.38 \Tstrut\Bstrut\\
        \texttt{[R]}  & \cellcolor{blue!25}51.80 & 67.53 & \cellcolor{blue!25}38.51 & 39.53 \Tstrut\Bstrut\\
        \texttt{[C + R]} & \cellcolor{blue!25}55.80 & 68.66 & \cellcolor{blue!25}53.59 & 38.52 \Tstrut\Bstrut\\        
        \hline
    	\multicolumn{5}{c}{\textit{\textbf{Linear Mapping}} (\textit{t} $\rightarrow$ \textit{s})}\Tstrut\Bstrut\\
        \hline
        \texttt{[M]} & 53.66 & \cellcolor{blue!25}65.20 & 21.86 & \cellcolor{blue!25}16.54 \Tstrut\Bstrut\\
        \texttt{[C]} & 54.93 & \cellcolor{blue!25}66.46 & 52.15 & \cellcolor{blue!25}40.38 \Tstrut\Bstrut\\
        \texttt{[R]} & 52.13 & \cellcolor{blue!25}67.93 & 38.58 & \cellcolor{blue!25}39.49 \Tstrut\Bstrut\\
        \texttt{[C + R]} & 55.60 & \cellcolor{blue!25}69.73 & 53.52 & \cellcolor{blue!25}38.73 \Tstrut\Bstrut\\
        \hline
        \multicolumn{5}{c}{\textit{\textbf{1-Hidden Layer Feedforward Network}} (\textit{s} $\rightarrow$ \textit{t})}\Tstrut\Bstrut\\
        \hline
        \texttt{[M]} & \cellcolor{blue!25}51.40 & 63.66 & \cellcolor{blue!25}20.28 & 12.19 \Tstrut\Bstrut\\
        \texttt{[C]} & \cellcolor{blue!25}55.00 & 65.86  & \cellcolor{blue!25}52.84 & 40.38 \Tstrut\Bstrut\\
        \texttt{[R]} & \cellcolor{blue!25}52.26 & 68.80 & \cellcolor{blue!25}49.96 & 47.48 \Tstrut\Bstrut\\
        \texttt{[C + R]} & \cellcolor{blue!25}\textbf{57.20} & \textbf{70.20} & \cellcolor{blue!25}\textbf{54.07} & 46.51\Tstrut\Bstrut\\
        \hline
    	\multicolumn{5}{c}{\textit{\textbf{1-Hidden Layer Feedforward Network}} (\textit{t} $\rightarrow$ \textit{s})}\Tstrut\Bstrut\\
        \hline
        \texttt{[M]} & 49.80 & \cellcolor{blue!25}64.26 & 17.95 & \cellcolor{blue!25}18.26 \Tstrut\Bstrut\\
        \texttt{[C]} & 54.06 & \cellcolor{blue!25}66.00 & 53.80 & \cellcolor{blue!25}40.45 \Tstrut\Bstrut\\
        \texttt{[R]} & 52.46 & \cellcolor{blue!25}68.60 & 49.48 & \cellcolor{blue!25}47.07 \Tstrut\Bstrut\\
        \texttt{[C + R]} & 56.93 & \cellcolor{blue!25}70.06 & 54.28 & \cellcolor{blue!25}\textbf{47.07}\Tstrut\Bstrut\\
        \bottomrule
    \end{tabular}
  }    
  \caption{\textbf{Precision @ 1 of BDMA with different losses.} An ablation study of the impact of different loss combinations while training a model with BDMA. \texttt{[M]} = MSE, \texttt{[C]} = cosine, \texttt{[R]}  = RCSLS and \texttt{[C + R]} = cosine + RCSLS loss.}
  \label{tab:exp_base_lr_model_performance_ablation}
\end{table}

\paragraph{Embeddings \& Baselines.}
We use normalized and mean-centered FastText embeddings \cite{joulin2016fasttext}, learned from language-specific Wikipedia. We train two types of translation models: (a) a linear mapping with a weight matrix $W\in\mathbb{R}^{d\times d}$ for a $d$-dimensional embedding, and (b) a $1$ hidden layer feed forward network. For baseline comparisons, we retrain VECMAP \cite{artetxe2016learning,artetxe2018generalizing}, GeoMM \cite{jawanpuria2019learning} and RCSLS \cite{joulin2018loss}. When possible, we compare with BLISS(R) \cite{patra2019bilingual}, Joint Align \cite{wang2019cross}, Cross-lingual Anchoring \cite{ormazabal2020beyond} and LNMAP \cite{mohiuddin2020lnmap} using results previously reported for high resource languages. We train BDMA with a combination of cosine (\texttt{C}) and RCSLS (\texttt{R}) losses, and separate baseline methods for each language and translation direction pair.

\subsection{Impact of Polysemy}
In Table \ref{tab:exp_base_hr_model_performance_combined}, we observe  BDMA's performance translating words in high resource languages. BDMA's performance is better or equivalent in comparison to other methods. Additionally, we note that the translation model is trained with $5000$ unique pairs while \textit{Joint Align} \cite{wang2019cross} and \textit{cross-lingual anchoring} \cite{ormazabal2020beyond} are trained with the full MUSE training dataset for any given language pair which is greater than $5$K.\footnote{See Appendix \ref{sec:appendix_dataset} for the original dataset sizes.} Similarly, Table \ref{tab:exp_base_lr_model_performance_combined} shows the performance of different models on low resource languages compared to BDMA. BDMA with $1$-H FFN performs better than a linear mapping with an overall increase as high as $2.82\%$ while translating Japanese to English. The exception is for Hindi, where the performance drops by $3.8$\% (Hi $\rightarrow$ En). We see that the model benefits from bidirectional training when there are polysemous words in the evaluation corpus, improving the network's ability to generalize.

\subsection{Impact of Unique Vocabulary}

Similar to the previous experiment, we analyze the impact of BDMA with an evaluation dataset of unique pairs for both high resource and low resource languages. In contrast to Table \ref{tab:exp_base_hr_model_performance_combined}, Table \ref{tab:exp_base_hr_model_performance} shows that both linear mapping and $1$-H neural network are comparable to other baselines (except RCSLS) when there are no polysemous words. Adding additional layers to the network does not provide any benefit, which is consistent with findings from \citet{sogaard2018limitations} and \citet{ruder2019survey} that a linear mapping performs well for these language pairs. Table \ref{tab:exp_base_lr_model_performance_default} details experiments for the same under low resource language conditions. Although BDMA performs better for En $\rightarrow$ Ru and En $\rightarrow$ Ja, Hi $\rightarrow$ En continues to perform poorly. In contrast, its performance is comparable for \textit{Portuguese} where the reduction is $1.13$\% (En $\rightarrow$ Pt) only. 

Therefore, the $2$ main benefits of BDMA are: (a) it creates a single bidirectional word translation model while keeping the performance of the model comparable to baseline, and (b) the $1$-H FFN is a single network in comparison to LNMAP (which has $3$), while Linear BDMA has the same number of parameters as all other methods in Table \ref{tab:exp_base_hr_model_performance_combined} and \ref{tab:exp_base_lr_model_performance_combined}.

\subsection{Importance of Training Direction}
If the filtered training pairs do not contain polysemous words, why is the training direction important? This is because when the model is trained for a number of epochs, its optimal savepoint is chosen based on the forward translation performance for the given language pair direction. As seen in Table \ref{tab:exp_base_lr_model_performance_combined} and \ref{tab:exp_base_lr_model_performance_default}, the direction chosen to start model training can have an impact of forward and reverse translation performance. For example, the model training with \textit{Ru} $\rightarrow$ \textit{En} performs better than \textit{En} $\rightarrow$ \textit{Ru}.

\paragraph{Ablation Study.}
In Table \ref{tab:exp_base_lr_model_performance_ablation}, we assess the impact of using (combinations of) MSE, cosine and RCSLS distance functions $\mathcal{D}$. A combined cosine and RCSLS loss (\texttt{[C + R]}) performs the best and provides consistent forward ($s \leftarrow t$) and reverse translation ($t \leftarrow s$) performance (within $0.5\%$).

\section{Related Work}
Over the years, many supervised methods have been proposed. \citet{irvine-callison-burch-2013-supervised} learn a binary classifier for a language pair that predicts if a given word pair is a translation of each other or not. \citet{artetxe2016learning} implement Procrustes alignment while normalizing and mean centering word embeddings. \citet{xing2015normalized} add an orthogonal loss while aligning manifolds. In \citet{artetxe2018generalizing}, additional pre- and post-processing steps are provided. \citet{conneau2017word} propose a new retrieval method called cross-domain similarity local scaling (CSLS) in order to reduce the ``hubness'' problem. \citet{joulin2018loss} convert CSLS into a loss objective in order to optimize the translation matrix. An important challenge with linear mapping is that it assumes that source and target languages have a similar manifold structure; \citet{sogaard2018limitations} show this assumption is not true for many language pairs. \citet{nakashole2018characterizing} show that transformations need to be non-linear and are dependent on the word's local neighborhood. Instead of learning a mapping between languages separately, \citet{wang2019cross} jointly learn the monolingual and cross-lingual embeddings for the given language pair. \citet{ormazabal2020beyond} extend skip-gram to project source embeddings into a fixed target space and using them as anchors to iteratively learn the mapping.

\textbf{Cyclic Loss for Reverse Translation.} \citet{xu-etal-2018-unsupervised} perform unsupervised word alignment using the cycle consistency loss while computing the sinkhorn distance between a forward and reverse translation network. \citet{mohiuddin2019revisiting} train a dual autoencoder-discriminator architecture and use a cyclic loss to train a bi-directional model. LNMAP \cite{mohiuddin2020lnmap} extends the autoencoder architecture with a 2 layer mapping to learn a non-isomorphic mapping between languages. Our work differs as we reduce the number of parameters in the model (as it contains the mapping only) while training an \textbf{invertible network} that can perform both forward and back translation.

\section{Conclusion} 
We show how a non-linear mapping (invertible neural network) can be trained with a cyclic consistency loss, showing that a common isomorphic assumption is not strictly necessary \cite{sogaard2018limitations}. The network trained has fewer parameters in comparison to \citet{mohiuddin2020lnmap} while providing equivalent or improved performance on the low-resource word translation task.

\section*{Acknowledgements}
We thank our anonymous reviewers for their constructive reviews and the UMBC HPCF facility for computational resources. %
This work is in part supported by AFRL, DARPA, for the KAIROS program under agreement number FA8750-19-2-1003. The U.S. Government is authorized to reproduce and distribute reprints for Governmental purposes, notwithstanding any copyright notation thereon. The views and conclusions contained herein are those of the authors and should not be interpreted as necessarily representing the official policies or endorsements, either express or implied, of the Air Force Research Laboratory (AFRL), DARPA, or the U.S. Government.

\bibliography{references/gr_image.bib,references/manifold_alignment.bib,references/word_embeddings.bib,references/general.bib}{}
\bibliographystyle{acl_natbib}

\clearpage
\newpage
\appendix
\setcounter{table}{0}
\setcounter{figure}{0}
\renewcommand{\thetable}{A\arabic{table}}
\renewcommand{\thefigure}{A\arabic{figure}}

\section*{Appendix}
In the following sections, we provide information about hyperparameter values for each network architecture, statistics about the dataset and results from additional experiments. The experiments are conducted on a NVIDIA K20 GPU with $\approx 4$GB of RAM and NVIDIA V100 GPUs with $16$GB of RAM. Each model is trained on a single GPU. Linear models can be trained on K20s and the larger $1$-H-FFN are optimized on V100s.

\section{Hyperparameters}
Following are the hyper-parameters used in our experiments:
\begin{table}[h]
    \centering
    \begin{tabular}{c|c}
        \toprule
        \textbf{Hyper-parameter} &  \textbf{Value} \\
        \hline
        \texttt{batch size} & $128$ \\
        \texttt{lr_decay} & $0.98$ \\
        \texttt{lr_shrink} & $0.5$ \\
        \texttt{map_beta} & $0.001$ \\
        \texttt{max_vocab} & $200000$ \\
        \bottomrule
    \end{tabular}
    \caption{Hyperparameters for BDMA experiments.}
    \label{tab:app_hyp}
\end{table}

As seen in table \ref{tab:app_hyp}, the maximum vocabulary (\texttt{max_vocab}) size is $200K$. The vocabulary is selected by taking $200K$ words that have the highest frequency. \texttt{map_beta} is the parameter that controls the contribution of the orthogonal loss to the overall loss function. The network is trained with an Adam optimizer \cite{kingma2014adam} having a learning rate of $0.0005$. The word embeddings are preprocessed i.e. they are normalized and centered. The $1$ hidden layer feedforward network used to perform alignment has a hidden layer size of $4096$. The activation function of the hidden layer  is \texttt{tanh}.

\section{CCL Correlation with Linear Mapping}
As observed in equation \ref{eqn:l_mse}, a linear relationship between source and target language embeddings can be learned by minimizing the squared loss between them. Although, in practice, an additional orthogonal constraint $\mathcal{L}_{ortho} = WW^T - I$ is added \cite{xing2015normalized} as shown in the equation below:
\begin{equation}
    \begin{aligned}
    \mathcal{L}_{\textrm{mse}} = 
    \sum_{i \in V^p}\overbrace{\left\|f_a(m^s_i) - m^t_i\right\|^2_2}^{\mathcal{L}_{\textrm{mse}}^i} +
    (WW^T - I),
    \label{eqn:l_mse_ort}
    \end{aligned}
\end{equation}

Minimizing $\mathcal{L}_{ortho}$ makes the linear mapping \textit{implicitly} bidirectional able to map words from the target to source language. In comparison, $\mathcal{L}_{ccl}$ in equation \ref{eqn:l_ccl} trains a non-linear neural network or linear mapping to be \textbf{explicitly} bidirectional. Thus $\mathcal{L}_{ccl}$ can be considered as an extension of $\mathcal{L}_{ortho}$.

\begin{table}[h!]
    \centering
    \begin{tabular}{c|c|c}
        \toprule
        \textbf{Target Language} & \textbf{Train} & \textbf{Test}\Tstrut\Bstrut\\
        \hline
        French & $10872$ & $2943$ \Tstrut\Bstrut\\
        German & $14677$ & $3660$ \Tstrut\Bstrut\\
        Italian & $9657$ & $2585$ \Tstrut\Bstrut\\
        Spanish & $11977$ & $2975$ \Tstrut\Bstrut\\
        \hline
        Russian & $10887$ & $2447$ \Tstrut\Bstrut\\
        Hindi & $8704$ & $2032$ \Tstrut\Bstrut\\
        Japanese & $7135$ & $1799$ \Tstrut\Bstrut\\
        Portuguese & $11185$ & $2827$ \Tstrut\Bstrut\\
        \bottomrule
    \end{tabular}
    \caption{\textbf{MUSE Dictionary Size}. The table shows the \textit{target} language, the number of pairs in the training and pairs present in the test dictionary where the \textit{source} language is \textit{English}.}
    \label{tab:exp_muse_dataset_size_tgt}
\end{table}

\begin{table}[h!]
    \centering
    \begin{tabular}{c|c|c}
        \toprule
        \textbf{Source Language} & \textbf{Train} & \textbf{Test}\Tstrut\Bstrut\\
        \hline
        French & $8270$ & $2342$ \Tstrut\Bstrut\\
        German & $10866$ & $2827$ \Tstrut\Bstrut\\
        Italian & $7364$ & $2102$ \Tstrut\Bstrut\\
        Spanish & $8667$ & $2416$ \Tstrut\Bstrut\\
        \hline
        Russian & $7452$ & $2069$ \Tstrut\Bstrut\\
        Hindi & $8001$ & $1963$ \Tstrut\Bstrut\\
        Japanese & $6819$ & $1952$ \Tstrut\Bstrut\\
        Portuguese & $7582$ & $2148$ \Tstrut\Bstrut\\
        \bottomrule
    \end{tabular}
    \caption{The table shows the \textit{source} language, the number of pairs in the training and pairs present in the test dictionary where the \textit{target} language is \textit{English}. }
    \label{tab:exp_muse_dataset_size_src}
\end{table}

\section{Dataset}
\label{sec:appendix_dataset}
\textbf{MUSE} \cite{conneau2017word}. As described in \S \ref{subsec:exp_muse_dataset}, the dataset has $110$ bilingual dictionaries and contains pairs with \textit{English} being the source or target language. Additionally, \textit{Non-English} language pairs are available for European languages that includes \textit{German}, \textit{Spanish}, \textit{French}, \textit{Italian} and \textit{Portuguese}. Each bilingual dataset has a vocabulary of $5000$ unique source language words to train the translation model and $1500$ unique words to evaluate them. Because the pairs are not unique and contain polysemous source words (the target word is always unique), the overall size of training and test dictionaries is greater than $5000$ and $1500$.

Tables \ref{tab:exp_muse_dataset_size_tgt} and \ref{tab:exp_muse_dataset_size_src} show the dataset size from the original MUSE dataset. The tables show that samples for different language pairs contain polysemous words that expand dataset size by $36.8$\% to $123.7$\% in comparison to BDMA (in table \ref{tab:exp_base_lr_model_performance_combined}, \ref{tab:exp_base_lr_model_performance_default} and \ref{tab:exp_base_hr_model_performance_combined}) that is trained with $5000$ unique pairs only.

\section{Additional Loss}
\label{sec:app_additional_loss}
In \S \ref{sec:bi-loss}, we showcased how MSE is adapted for $\mathcal{L}_{\textrm{ccl}}$. Similarly, cosine and Relaxed CSLS loss can be modified for BDMA too. In an adapted version of cosine loss, we minimize the following:
\begin{equation*}
\small
\min_{\theta_f} \sum_{i \in V^p} (1 - |f_a(m^s_i) \cdot m^t_i|) + (1 - |f_b(m^t_i) \cdot m^s_i|)
\label{eqn:l_cosine}
\end{equation*}

In order to modify RCSLS \cite{joulin2018loss}, we first take look at CSLS \cite{conneau2017word} criteria for retrieval:
\begin{dmath}
\text{CSLS}(m^s_i, m^t_i) = -2\text{cos}(m^s_i, m^t_i) + \frac{1}{\text{k}} \sum_{m^t_j \in \mathcal{N}^t(\mathcal{W}\cdot m^s_i)}\text{cos}(\mathcal{W}m^s_i,m^t_j) + \frac{1}{\text{k}} \sum_{m^s_j \in \mathcal{N}^s(\mathcal{W}^{\mathcal{T}}\cdot m^t_i)}\text{cos}(m^s_j,\mathcal{W}^{\mathcal{T}}m^t_i))
\label{eqn:l_csls}
\end{dmath}
where $\mathcal{N}^s(x)$ is the neighborhood of $x$ in the source manifold and $\mathcal{N}^t(y)$ is the same in the target, $\text{k}$ is the number of nearest neighbors and $\mathcal{W}$ is assumed to be orthogonal. \citet{joulin2018loss} relax the cosine criteria in RCSLS i.e. $\text{cos}(\mathcal{W}m^s_i, m^t_i) = {m^s_i}^{\mathcal{T}}\mathcal{W}^{\mathcal{T}}m^t_i$. Hence RCSLS becomes:

\begin{dmath}
\text{RCSLS}(m^s_i, m^t_i) = -2{m^s_i}^{\mathcal{T}}\mathcal{W}^{\mathcal{T}}m^t_i + \frac{1}{\text{k}} \sum_{m^t_j \in \mathcal{N}^t(\mathcal{W}\cdot m^s_i)}{m^s_i}^{\mathcal{T}}\mathcal{W}^{\mathcal{T}}m^t_j + \frac{1}{\text{k}} \sum_{m^s_j \in \mathcal{N}^s(\mathcal{W}^{\mathcal{T}}\cdot m^t_i)}{m^s_j}^{\mathcal{T}}\mathcal{W}^{\mathcal{T}}m^t_i
\label{eqn:l_rcsls}
\end{dmath}

In BDMA, we replace the orthogonal matrix $\mathcal{W}$ with a mapping that is either linear or non-linear (neural network). RCSLS changes to:

\begin{dmath}
\text{RCSLS}(m^s_i, m^t_i) = -2f_a(m^s_i)m^t_i + \frac{1}{\text{k}} \sum_{m^t_j \in \mathcal{N}^t(f_a(m^s_i))}f_a(m^s_i)m^t_j + \frac{1}{\text{k}} \sum_{m^s_j \in \mathcal{N}^s(f_bm^t_i)}m^s_jf_b(m^t_i)
\label{eqn:l_rcsls_m}
\end{dmath}
In equation \ref{eqn:l_rcsls_m}, $f_a$ and $f_b$ are the forward and reverse flow projections of $m^s_i$ and $m^t_i$ respectively.

\end{document}